\documentclass[10pt,twocolumn,letterpaper]{article}

\usepackage[pagenumbers]{cvpr} 

\usepackage{graphicx}
\usepackage{caption}
\usepackage{amsmath}
\usepackage{amssymb}
\usepackage{booktabs}
\usepackage{float}
\usepackage{multirow}
\usepackage{makecell}
\usepackage{comment}
\usepackage{xcolor}
\usepackage{bbm}
\usepackage[normalem]{ulem}
\usepackage[accsupp]{axessibility}  
\newcolumntype{P}[1]{>{\centering\arraybackslash}p{#1}}
\newcommand*{\MyIndent}{\hspace*{0.5cm}}
\usepackage[pagebackref,breaklinks,colorlinks]{hyperref}

\usepackage[capitalize]{cleveref}
\crefname{section}{Sec.}{Secs.}
\Crefname{section}{Section}{Sections}
\Crefname{table}{Table}{Tables}
\crefname{table}{Tab.}{Tabs.}

\def\cvprPaperID{5} 
\def\confName{CVPR}
\def\confYear{2023}
\newcolumntype{P}[1]{>{\centering\arraybackslash}p{#1}}

\begin{document}

\title{Human Gesture and Gait Analysis for Autism Detection}
\author{Sania Zahan $^1$, Zulqarnain Gilani $^2$, Ghulam Mubashar Hassan $^1$ and Ajmal Mian $^1$\\
$^1$ The University of Western Australia
$^2$ Edith Cowan University\\
{\tt\small sania.zahan@research.uwa.edu.au} \hspace{1cm}
{\tt\small s.gilani@ecu.edu.au}\\
{\tt\small \{ghulam.hassan, ajmal.mian\}@uwa.edu.au}
}
\maketitle


\begin{abstract}   
   Autism diagnosis presents a major challenge due to the vast heterogeneity of the condition and the elusive nature of early detection. Atypical gait and gesture patterns are dominant behavioral characteristics of autism and can provide crucial insights for diagnosis. Furthermore, these data can be collected efficiently in a non-intrusive way, facilitating early intervention to optimize positive outcomes. Existing research mainly focuses on associating facial and eye-gaze features with autism. However, very few studies have investigated movement and gesture patterns which can reveal subtle variations and characteristics that are specific to autism. To address this gap, we present an analysis of gesture and gait activity in videos to identify children with autism and quantify the severity of their condition by regressing autism diagnostic observation schedule scores. Our proposed architecture addresses two key factors: (1) an effective feature representation to manifest irregular gesture patterns and (2) a two-stream co-learning framework to enable a comprehensive understanding of its relation to autism from diverse perspectives without explicitly using additional data modality. Experimental results demonstrate the efficacy of utilizing gesture and gait-activity videos for autism analysis.

\end{abstract}

\section{Introduction}
\label{sec:intro}
Autism Spectrum Disorder (ASD) is a neurodevelopmental condition that poses significant communication and behavioral challenges \cite{CDC_ASD}. Children with autism usually interact or behave differently compared to typically developing (TD) children. Their cognitive abilities can vary greatly, ranging from exceptional talent to severe challenges in learning, thinking, and problem-solving \cite{Karen_2022}. Whereas the reported prevalence of autism worldwide is one in 100 children, this figure is just an average, and the actual numbers can be substantially higher \cite{who2022}. 

Autism can be detected in children as early as 18 months old or even younger \cite{Catherine2006}. Early intervention and specialized support services can substantially improve a child’s development \cite{NRC2001}. However, delayed diagnosis often results in a missed opportunity for children to receive crucial early assistance \cite{Catherine2006}.  Furthermore, diagnosis can be challenging due to the lack of reliable and efficient diagnostic tools. Parents are often reluctant to accept the condition or fail to detect subtle behavioral cues \cite{Sudhinaraset_2013, GROGAN2023102102}. This can lead to months of wasted time before a child gets access to proper support.

Clinically, autism is diagnosed in a face-to-face interactive session with a trained health professional who analyzes certain behavioral traits using verbal and non-verbal tasks. Communication and language assessment are essential factors in the diagnosis \cite{Hudry_2023}. However, approximately 40 percent of children with autism are nonverbal \cite{autism_speaks_2022} which further complicates the diagnosis process for this specific population. An initial diagnosis in a more suitable and accessible way is needed
to facilitate higher detection accuracy of ASD. This can ensure early intervention and access to customized therapies for effective management. 

Throughout the years, researchers have proposed several methods for ASD detection \cite{Andrea_2018, Qandeel2019, Boutrus_2019, Boutrus_2021, Diana2021, Jing_2022_TC, NEGIN_2021, Tian_2019, Sabater_2021_CVPR, Marinoiu_2018, Ahmed_2021, Chen_2019}. Many of these methods primarily focus on appearance-based features \cite{Qandeel2019, Boutrus_2019, Jing_2020, Diana2021, Boutrus_2021, Jing_2022_TC}. However, 
appearance does not provide detailed insights into the autistic behavioral traits such as social-emotional exchanges, communication difficulties, stereotyped activities, etc  which form a crucial part of the diagnosis \cite{Healthdirect_2022}.
Recent research shows that children with autism usually exhibit distinctive gait and gesture activity patterns \cite{autism_speaks_2022}. Leveraging these patterns can facilitate the extraction of a distinguishable feature distribution, thus improving classification accuracy.

The unique atypical gesture activities of children with autism may include:
\begin{itemize}
    \item \textbf{Repetitive movements} such as rocking, arm flapping, or spinning, known as ``stereotypy".
    \item \textbf{Limited range of gestures} and may also have difficulty understanding the gestures of others.    
    \item \textbf{Atypical gait and posture} resulting in unbalanced movement and instability in joints.
    \item \textbf{Impaired motor coordination} which can lead to difficulties with fine motor skills, such as writing, or gross motor skills, such as running.    
\end{itemize}



In this paper, we propose a video-based method that takes a holistic view of gait and gesture to detect subtle disparities in ASD children. We evaluate the proposed method on a video dataset collected from children with and without ASD. Experimental results demonstrate the effectiveness of gesture activities in accurately identifying children with autism. This has the potential to significantly impact early diagnosis and management of ASD by providing a reliable, non-intrusive, and efficient tool for autism classification. Moreover, the action perspective facilitates the assessment of children with limited verbal communication. 

The severity of autism is measured using Autism Diagnostic Observation Schedule (ADOS) score. However, there is currently no research on autism severity prediction in terms of ADOS score regression. Since the severity of autism influences gait and gesture patterns, a comprehensive analysis of activity videos can help identify subtle irregularities to assess the severity. Therefore, in this paper, we represent our analysis of ADOS score regression.

In a nutshell, our contributions are as follows. 
\begin{itemize}
    \item We propose a novel angular feature matrix which is embedded into the input skeleton and encoded using a Graph Convolutional Network (GCN). Our proposed angle embeddings enable the GCN to detect the peculiar slant in the gait posture of ASD children. To the best of our knowledge, this is the first research that focuses on this aspect of gait posture in the machine learning paradigm.

    \item We automatically predict ADOS scores which are highly correlated with ADOS scores measured by human experts. 

    \item We perform a detailed analysis of the gait posture of both ASD and TD children and investigate the asymmetry in their gait. 
\end{itemize}

\begin{figure*}[!hbtp]
\centering
\includegraphics[width=1.0\textwidth]{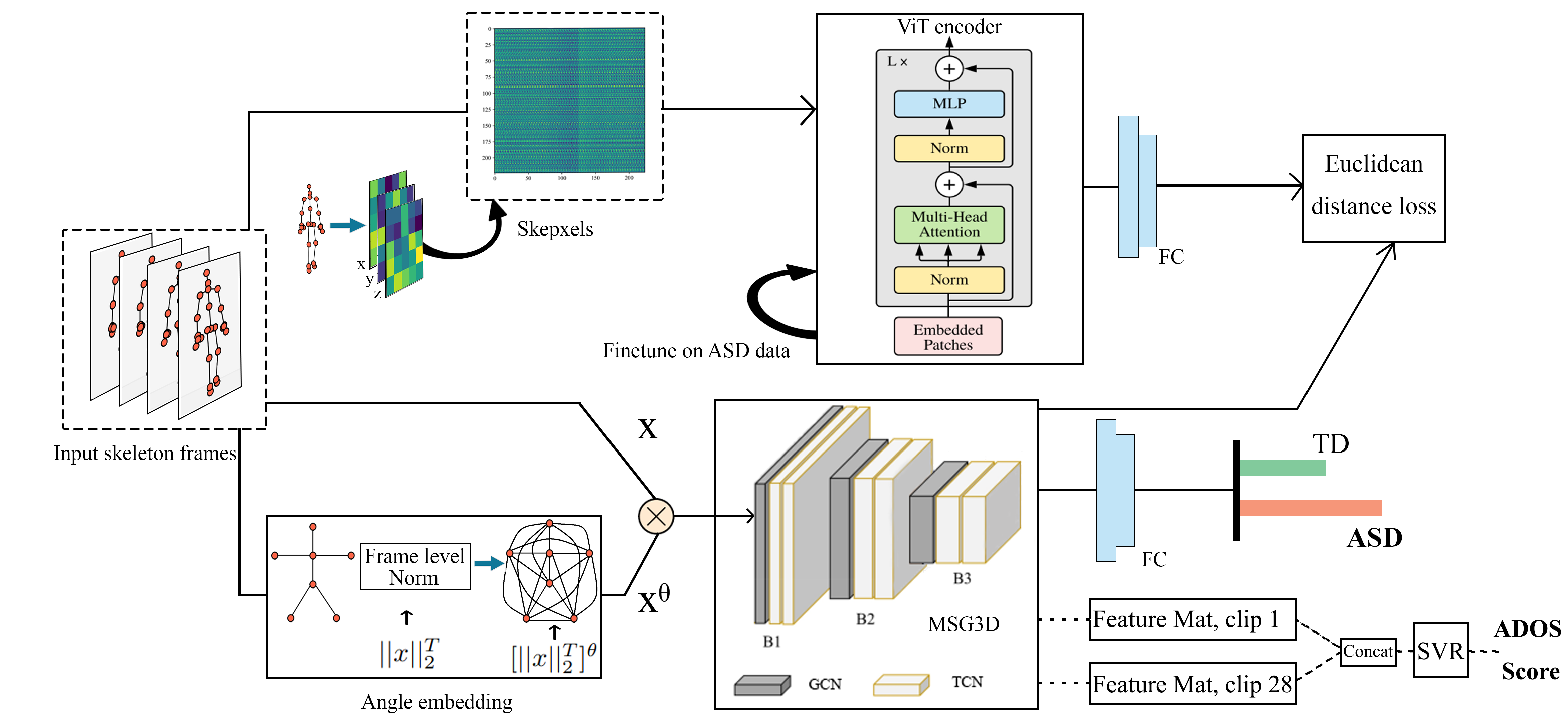}
\vspace{-0.4em}
\caption{Proposed method: Angular feature matrix $[||x||_2^T]^\theta$ is embedded into the input skeleton ($||x||_2^T$ indicates normalization), which is then encoded by MSG3D using GCN followed by TCN. Vision Transformer (ViT) computes an aggregated embedding of the Skepxels and is used during training for pair-wise distance loss. For classification, an FC layer is used after the GCN layer. For score regression, clip-based features are extracted from the GCN and concatenated. SVR is used to predict the final ADOS scores at the video level.} 
\label{model}
\vspace{-1em}
\end{figure*}  


\section{Related Work}\label{lit_rev}               
Existing research has explored various approaches for the detection of autism, with a prominent emphasis on facial expression and eye-gaze pattern-based techniques. 

\subsection{Facial expression and eye-gaze pattern}
Physical appearance is a distinguishable feature of autism. In \cite{Qandeel2019}, developmental delays are detected from physical appearance in home videos. Asymmetry in facial appearance is studied in \cite{Diana2021}. The study finds that asymmetric features are more common in people with a history of ASD. Other studies also corroborate that children with ASD display higher facial asymmetry \cite{Boutrus_2019, Boutrus_2021}.

The eye-gaze pattern is also a salient marker of autism as ASD children show decreased attention than TD children \cite{riby_2009}. Their facial expressions and eye gaze lack engagement with surrounding environments. This reduced eye-gaze pattern is stable across all ages and cultures \cite{ma_2021}. The stacked accumulative histogram proposed in \cite{Jing_2020}, captures these anomalies in eye movement trajectory. The method requires manual labeling of the eye region, and a tracking algorithm \cite{Kalal_2012} to obtain the trajectories. The displacement features represent higher disparity among different visual zones for children with ASD. AttentionGazeNet \cite{Jing_2022_TC} generates a projection of screen coordinates from 3D gaze vectors. Experiments indicate that gaze vectors are more dispersed for children with ASD.  Similarly, \cite{Jiang_2017} also finds a substantial difference in eye movement patterns between ASD and TD children.

\subsection{Gesture pattern}
In \cite{Anzulewicz_2016}, a wide disparity is found in hand gesture patterns between ASD and TD children. During gameplay on a smart tablet, children with ASD used more force and gesture pressure within a greater mean area. 
Another gesture-based research in \cite{Andrea_2018} hypothesizes that the disparity in gesture patterns in performing actions also extends to the onset, embedding information about the intention. Thus, the intended gestures can be used to diagnose children with ASD. These studies indicate the usability of motor functions in ASD analysis.


Another research direction focuses on classifying atypical actions from videos. The Bag-of-visual-words approach \cite{NEGIN_2021} treats image grids as visual words to recognize relevant feature descriptors. In \cite{Tian_2019}, a temporal pyramid network is used to create layers of feature maps from videos with extended duration. A separate repetitive behavior discriminator is used to boost the training process by distinguishing samples with atypical actions. The atypical action classification approach in \cite{Sabater_2021_CVPR} uses an anchor action instance to facilitate pair-wise similarity with the target embedding. \cite{Marinoiu_2018} analyzes action and emotion recognition from ASD therapy videos. 

In \cite{Ahmed_2021}, skeleton-based handcrafted features are used to classify ASD children. The attention-based ASD screening method in \cite{Chen_2019} exploits multiple modalities to embed complementary multimodal knowledge in a shared space. 

Though extensive research has been done on appearance-related abnormalities in autism, this approach offers a limited perspective. Gesture pattern, on the other hand, provides a comprehensive perspective on physical behavior. However, existing gesture-based autism detection methods mainly use end-to-end deep learning and do not incorporate attention to the underlying mechanisms of atypical behaviors in ASD children. Therefore, in this study, we investigate atypical gesture patterns and integrate them into the learning process to enhance the representation of anomalies.


\section{Method}\label{Methodology}
Our proposed method is built on the hypothesis that ASD is distinguishable solely from gesture patterns \cite{Marchena_2019, Fourie_2020}. Skeleton-based representation facilitates visualizing gesture patterns effectively, which can be encoded into spatio-temporal embedding using graph convolution. By incorporating the joint angles, our method enhances feature representation and enables the extraction of comprehensive structural aspects and aberrations in human body movement. Overall, the skeleton frames are embedded with angle information, then encoded using MSG3D \cite{liu2020disentangling} and a fully connected layer (FC) generates the final mapping to the classes. For score regression, the output from MSG3D is concatenated and then used in SVR. Vision transformer is only used during training to expand the learning capacity by processing the data from a different perspective. Figure~\ref{model} illustrates our proposed method.

\subsection{Graph Convolutional Network: GCN}
Graph convolutional networks (GCN) \cite{stgcn2018aaai, liu2020disentangling} can capture the underlying structures and encode how nodes are connected. GCN applies a localized convolutional operation to each node and its neighbors. A GCN block incorporates separate spatial (GCN) and temporal convolution (TCN) to leverage frame-wise and global attention. GCN applies filtering over the spatial dimension to encode spatial features and TCN encodes temporal features by applying filters over the temporal dimension. Multiple stacked GCN blocks generate increasingly abstract representations of the input graph.

Human skeletons can be represented by graphs as $G = (V, E)$ where $V = \{v_1, ... v_N\}$ is the set of nodes (joints) and $E = \{e_1, ..e_N\}$ is the set of edges (bones). The adjacency matrix $A \in R^{N \times N}$ represents local joint adjacency. We used MSG3D as our GCN encoder \cite{liu2020disentangling} to aggregate important spatio-temporal features. MSG3D for skeleton graph convolution can be represented below, as mentioned in Eq 1 in \cite{liu2020disentangling}

\begin{equation}
    X_t^{(l+1)} = \sigma(\tilde{D}^{-\frac{1}{2}}\tilde{A}\tilde{D}^{-\frac{1}{2}}X_t^{(l)}\theta^{(l)}),
\end{equation}
where $\tilde{A}$ is the modified adjacency matrix, $\tilde{D}$ is the diagonal degree matrix of $\tilde{A}$, $\theta$ is the learnable weight matrix, and $\sigma$ is the ReLU activation function. $\tilde{A}$ is modified $A$ with added self-loops $I$ and is computed by identifying the shortest $k$ distance joints and subtracting graph powers as $\tilde{A}_{(k)} = I + \mathbbm{1} (\tilde{A}^k \geq 1) - \mathbbm{1}(\tilde{A}^{k-1} \geq 1)$. MSG3D uses multiple graph convolutions at different scales to extract different levels of details or resolution. Thus, it is able to efficiently capture both local and global patterns in the input graph. The encoded features are then used for ASD classification and ADOS score prediction.

\subsection{Angle embedding}
\begin{figure}[t]
\centering
\includegraphics[width=\columnwidth, height=3.8cm]{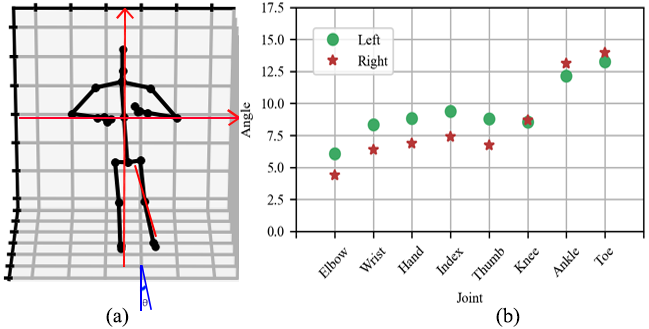}
\vspace{-1.9em}
\caption{(a) Initial gait pose of an ASD skeleton: the red lines indicate the coordinates through the spine joint. (b) Calculated angles of the left (green) and right (red) side joints.} 
\label{angle_embed_skel}
\vspace{-0.3cm}
\end{figure}

Our statistical analysis of variations and joint distributions in gait between ASD and TD children (detail in Section~\ref{stat_analysis}) revealed that ASD children have a tendency to walk with a slanted gait posture compared to TD children. Additionally, we also found that atypicalities and asymmetry on the left and right sides of their body create irregular joint positions and movements. Due to slanted and asymmetric gait, the joints in the ASD skeleton samples form a much higher angle with the spine line. We find this to be a distinctive feature of ASD children. Figure~\ref{angle_embed_skel} illustrates the slanted posture and calculated joint angles. 

We embed the angle features into the input skeleton stream to enhance the inherent gait disorder prevalent in children with ASD. First, the input skeletons are normalized over the frame dimension. Then angle between each joint is calculated which creates a $25\times25$ feature matrix. Finally, this angle matrix is multiplied with the input skeleton stream over the joint dimension to generate the embedded features. Eq~\ref{angle_eq} illustrates the computation process.

\begin{equation}\label{angle_eq}
\begin{split}
X_{norm} = \sqrt{\sum_{t=1}^{T}|X|^2}; \quad \bar{X} = \frac{X}{X_{norm}}; \quad X^\theta_{i,j} = \bar{X}_i \cdot \bar{X}_j
\end{split}
\end{equation}
where $X$ is the input skeleton,  $\bar{X}$ indicates L2 normalization over frame dimension $T$ and $X^\theta_{i,j}$ represents the cosine angle of each  joint $i$ with all joints $j$ where $j=25$,  calculated using dot product of each joint with all 25 joints as 
$\bar{X}_i \cdot \bar{X}_j = |\bar{X}_i| \cdot |\bar{X}_j|cos\theta$. This embedding process increases the distinction between ASD and TD skeleton feature space.

\subsection{Skeleton Picture Elements: Skepxel}
\begin{figure}[t]
\centering
\includegraphics[width=7.5cm]{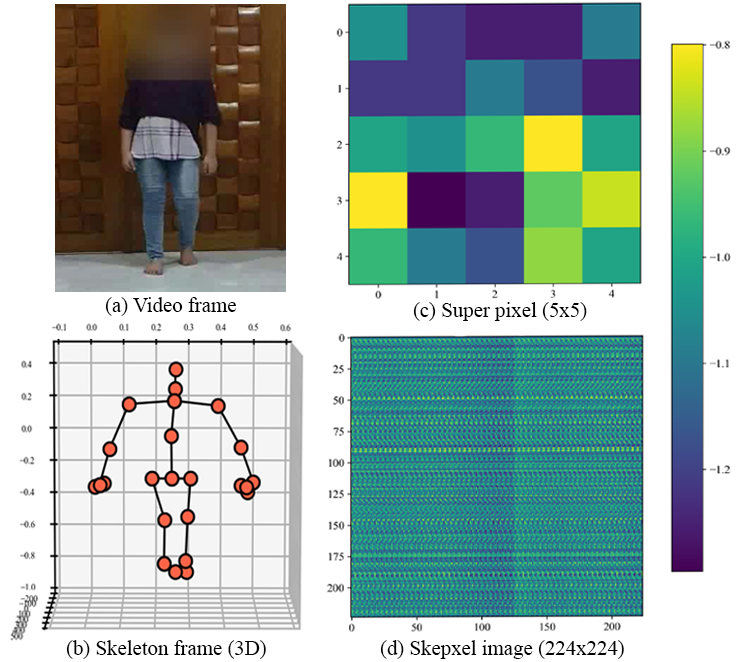}
\vspace{-0.5em}
\caption{Illustration of (a) a video frame, (b) skeleton frame, (c) the corresponding 5x5 super pixel, and the (d) Skepxel image \cite{Liu_2019_CVPR_Workshops}.} 
\label{gait_sample}
\vspace{-0.4cm}
\end{figure}

Mainstream vision models such as vision transformers (ViT) have exhibited exceptional performance, yielding remarkable results in various tasks. However, our experimental results indicate that direct use of skeleton frames in ViT produces poor performance since the joint locations are not images. To address this limitation, we consider other representations of the skeleton frames. However, due to the limited number of joints available, the potential for generating diverse representations is constrained. The image representation of skeleton frames proposed in \cite{Du_2015, Liu_2019_CVPR_Workshops} offers a comprehensive view and is suitable for processing with ViT. However, \cite{Du_2015} focuses solely on global spatio-temporal information resulting in smaller resolution images. On the other hand, \cite{Liu_2019_CVPR_Workshops} takes into account both local and global spatio-temporal correlations of joints simultaneously, providing a more holistic representation of the data with higher resolution, termed as Skepxels. We leverage Skepxels to extract additional meaning from skeleton joints during training. The use of Skepxels allows us to effectively guide GCN during training to encode the complex motions and structures of different gesture patterns, leading to improved accuracy in ASD classification. Figure~\ref{gait_sample} represents a sample skeleton frame and corresponding Skepxel image.


Skepxels are constructed \cite{Liu_2019_CVPR_Workshops} by organizing the skeleton joints in different orders. Three channels of the joints act as the three RGB color channels. Super pixels in a single column represent different ordered joints of the same frame and each row contains superpixels from different temporal frames. This systematic construction results in an aggregated image representation that encodes heterogeneous semantic perceptions. For further details read the paper \cite{Liu_2019_CVPR_Workshops}.
 
\subsection{Vision transformer encoder: ViT}
As outlined in Figure~\ref{model}, we use ViT \cite{dosovitskiy2021an} to encode aggregated feature representation of Skepxels. ViT takes non-overlapping patches of the image and projects them onto a feature vector through a learnable linear projection. It uses positional embedding to retain the order of the patches. The self-attention mechanism allows it to selectively attend to relevant parts of the image, enabling it to capture global context more effectively than conventional CNNs. We use an MLP layer after ViT to map the Skepxel embedding to the same feature space as the GCN stream.


During the training phase, the Skepxel embedding is incorporated into the model training to facilitate co-learning of the spatial and temporal features. Euclidean distance loss between the Skepxel embedding and joint embedding from GCN is added with the classification loss. However, it is not used during the testing phase, where the model relies solely on the learned representations from the GCN stream.

\section{Datasets}\label{Dataset}
We evaluate our proposed method on two datasets. We use the Gait and Full Body Movement dataset for ASD classification \cite{Ahmed_2021} and the DREAM dataset for ADOS score regression \cite{billing_2020}.
\subsection{Gait and Full Body Movement dataset}
This dataset used Kinect v2 and Samsung note 9 to collect 3D joint coordinates (skeleton videos) and RGB videos \cite{Ahmed_2021}. The dataset was collected in a controlled environment where children walked 2.5m  (approximately two gait cycles) in front of the camera ten times. Then one gait cycle was extracted from an eligible candidate of the ten trials. Finally, faces were detected using Haar Cascade or MTCNN and blurred with a Gaussian filter to obscure the identities for anonymity in the RGB videos. The dataset contains fifty-nine children with ASD (9 of these children have severe ASD) and fifty TD children. There are a total of 109 samples. The dataset also contains seven augmented versions of each sample using jittering, scaling, left and right translation, horizontal and vertical flipping, and slicing, increasing the augmented dataset size to 700 samples. 

We follow two approaches for train and test sample selection: random shuffling of subjects, mentioned as random data and sliding window to select blocks of subject range, mentioned as block data in the results section. With block data sample selection, we ensure that each subject is evaluated at least once as either a training or testing sample. Furthermore, whenever a subject is selected for either training or testing, both original and augmented versions go to the same split to ensure no leakage of the augmented samples to the testing set or vice versa. 

\subsection{DREAM dataset}
The DREAM dataset incorporated 61 children (9 female) aged between 3 to 6 years with varied levels of autism and ADOS scores ranging from 7-20 \cite{billing_2020}. The samples were collected in a therapy environment where the children interacted with either a human therapist (SHT) or a humanoid robot (RET). They performed three different tasks: imitation, joint attention, and turn-taking. The sessions were conducted in the following manner. The interaction partner (human or robot) provided a discriminative stimulus and waited for the response from the child. Positive feedback or indication to repeat was provided depending on the child's behavior. Each session duration varied from 3 to 87 minutes, with a median duration of 32 minutes.

\begin{figure}[t]
\centering
\includegraphics[width=\columnwidth]{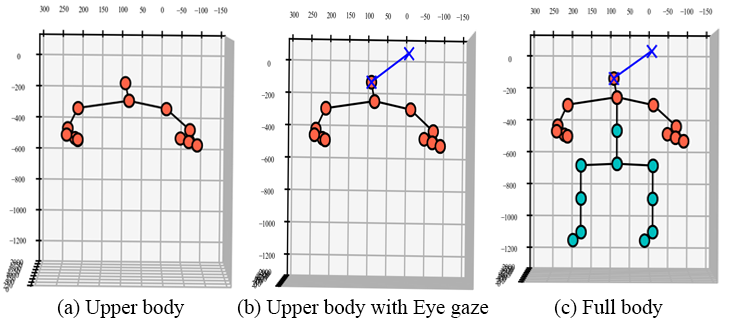}
\vspace{-2em}
\caption{Sample skeleton frame from DREAM dataset (a) original upper body skeleton from the dataset (b) upper body skeleton with eye gaze as additional head joint and (c) full body skeleton with interpolated lower joints.} 
\label{dream_dataset_samples}
\vspace{-1em}
\end{figure}

The dataset provides ADOS scores from two diagnosis sessions, initial and intervention. We consider only the initial diagnosis samples for this work as final ADOS scores are not published for the intervention sessions. ADOS score is a semi-structured autism diagnosis technique that is most commonly used to measure the severity of ASD. In addition to a raw total score, it includes individual scores for several skill factors such as communication, age, language level, interaction, etc., to categorize children into homogeneous groups \cite{Katherine2009}. Based on their age and the ADOS score evaluation module used, children with ASD can be divided into three classes: NonSpectrum (NS), Autism Spectrum Disorder (ASD), and Autistic (AUT, the most severe case of autism), following the metrics provided in \cite{Katherine2009}. We design our work to predict the ADOS score and classification of the samples into the above three classes based on the predicted scores. Figure~\ref{dream_dataset_samples} shows a sample skeleton.

\subsection{Data preprocessing}\label{Data_Preprocessing}
We redesign the skeletons in the DREAM dataset as it contains only upper-body joints. The missing joints are interpolated from the existing joints in the different direction. Thus the incomplete 10 joints skeleton structure is converted to a full-body structure with 25 joints. This enables us to process these skeletons using our proposed method where the GCN module expects the input to be 25 skeleton joints. We use the eye-gaze vector as the third head joint. This approach adds an extra perspective to our modified 25-joint skeletons by integrating the subject's visual attention. We replace any missing eye-gaze values with preceding values. The dataset is preprocessed to have a view-invariant transformation with the shoulder joints aligned with the x-axis and spine joints aligned with the z-axis. The spine joint is translated to the origin (0,0,0). We repeated frames where necessary to maintain a fixed-length video.

For the Gait and Full Body Movement dataset, we do not perform rotation as it will eliminate the slanted gait posture of ASD skeletons.
\section{Experiments}\label{Experiments}
Our statistical analysis of the gait pattern reveals discriminative atypicalities in ASD samples and our proposed method exploits this insight for autism classification and ADOS score regression.

\subsection{Statistical analysis}
In this section, we present a comparative statistical analysis between ASD and TD samples. ASD samples exhibited distinctive variations and asymmetry over the entire population.\\

\begin{figure}[t]
\centering
\includegraphics[height=5cm]{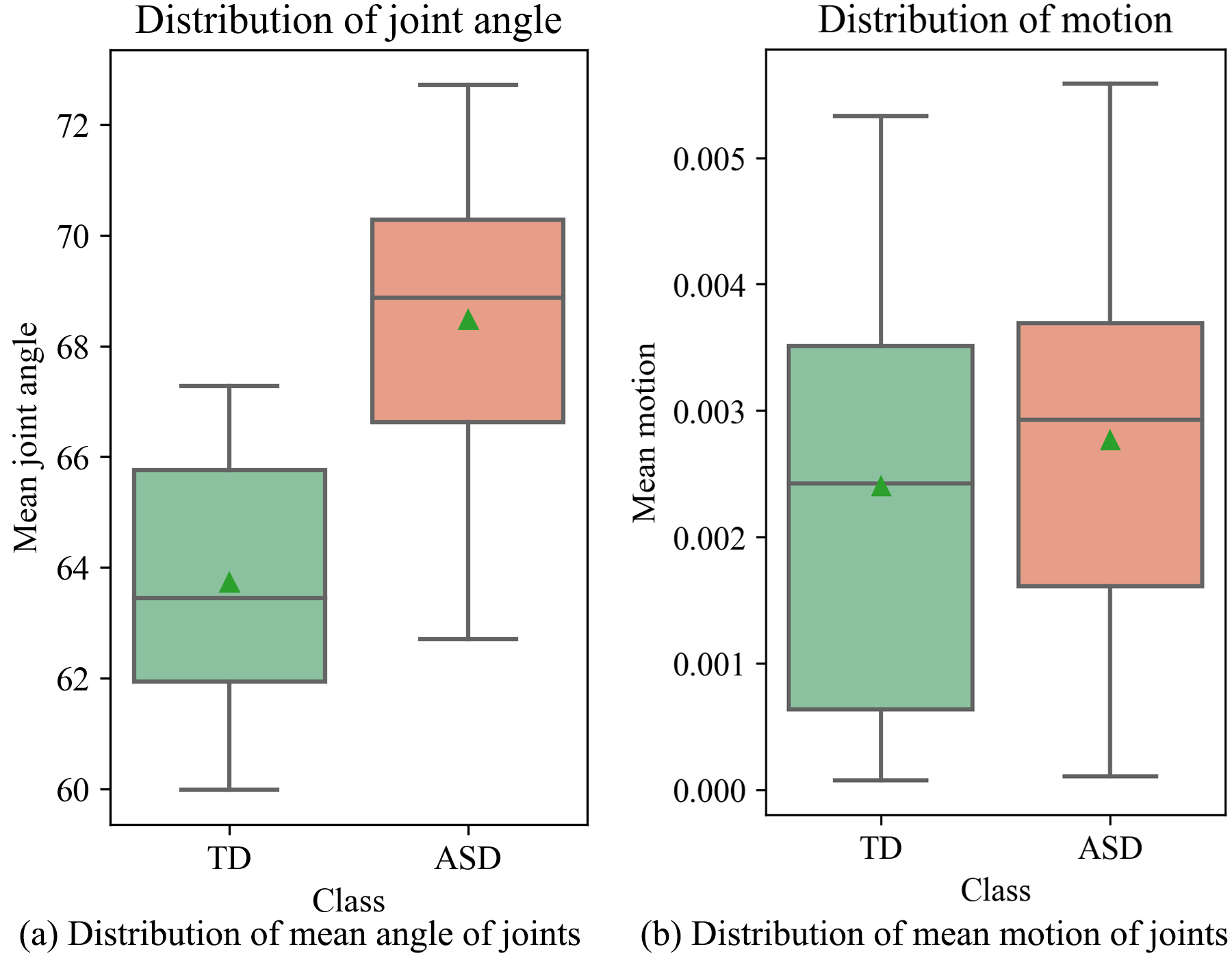}
\caption{(a) Distribution of joint angle, higher median indicates ASD children have a much higher median joint angle. (b) Distribution of motion of TD and ASD samples; a larger box of TD samples demonstrates that TD children express higher variations in motion while walking.} 
\label{boxplot_of_gait}
\vspace{-0.5cm}
\end{figure}

\textbf{Variation in joint and motion distribution:} \label{stat_analysis}
Children with ASD represent restricted and repetitive behavioral patterns \cite{Boutrus_2021}. They usually tend to have a slower gait and may have visible difficulty while walking. These atypical gaits create a slanted posture. We analyzed the angle of each joint with the spine (center) joint in each frame and calculate the mean. Figure~\ref{boxplot_of_gait} (a) illustrates the distribution of these joint angles for TD and ASD children. ASD children present a much higher distribution than TD children, demonstrating a higher mean joint angle. 

Figure~\ref{boxplot_of_gait} (b) shows the distribution of mean motion. The limited range of gestures and slower gait cycle of ASD children means that the distribution of their movement pattern will be less dispersed. On the other hand, TD children usually show comparatively more diverse motion, resulting in a broader motion distribution.

\MyIndent\textbf{Asymmetry in gait:}
Recent studies have revealed that children diagnosed with ASD exhibit hypermasculine traits and asymmetry in their facial morphology \cite{Diana_2017, Diana_2020, Diana_2022, Boutrus_2019, Diana2021, Boutrus_2021}. Based on the atypical gesture behaviors observed in children with ASD, we hypothesize that this asymmetry may also manifest in their walking patterns or gait. In order to ascertain whether there is an asymmetry in the gait, we compared the angle, motion, and distance between joints on the left and right sides of the body. Our observations are illustrated in Figure~\ref{assymetry_of_gait}. 

\noindent%
\begin{minipage}{\columnwidth}
\makebox[\columnwidth]{
  \includegraphics[keepaspectratio=true,scale=0.267]{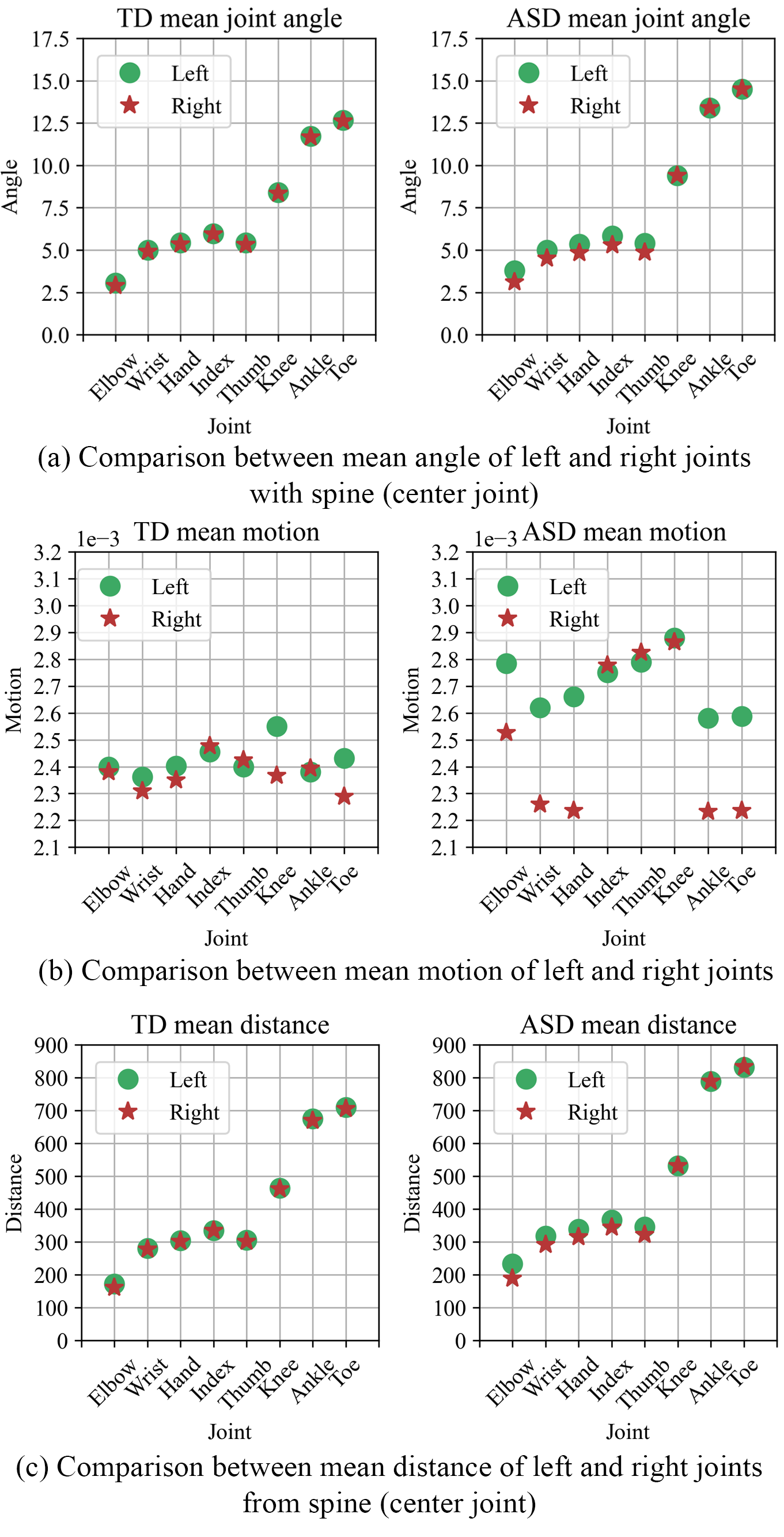}}
  \vspace{-2em}
\captionof{figure}{Illustration of asymmetry in the joints of the left and right sides of the body. (a) Comparison between mean joint angle (in degrees): left and right side joints are overlapped in TD samples whereas deviated in ASD samples. (b) Comparison between mean motion. ASD samples indicate much higher as well as asymmetric movement patterns in left and right side joints. (c) Mean distance (in centimeters) of the left and right side joints from the spine is depicted. Distance in ASD samples is more dispersed which represents that ASD children have asymmetric hand and leg positions during walking.}\label{assymetry_of_gait}
\end{minipage}

A typical person tends to have uniform hand and leg movement coordination in a complete gait cycle. A higher difference in angle, motion, and distance between the left and right side joints indicates higher asymmetry. We select 16 joints, including 10 hand joints (left and right) and 6 leg joints (left and right) and calculate the angle, motion, and distance of each joint with the corresponding spine joint in the same frame. As can be seen in Figure~\ref{assymetry_of_gait} (a), (b), and (c), ASD samples depict higher differences in the values between left and right joints than TD. Thus, we can conclude that the gait of children with ASD expresses higher asymmetry. 

\subsection{Quantitative results}\label{results}
Experimental results on the datasets for ASD detection and ADOS score regression are discussed in the following sections.\\

\textbf{Results on Gait and Full Body Movement dataset:} Table~\ref{CrossValidation_results} presents average results from 10-fold cross-validation along with standard deviation on the  Gait and Full Body Movement dataset. In the Table, \textbf{MSG3D} indicates results for the initial model proposed in \cite{liu2020disentangling}. We used pretrained weights of MSG3D trained on NTU RGB+D 120 action recognition dataset \cite{Liu_NTU120_2020}. \textbf{Angle infusion} represents joint angle embedding into the input skeletons, and \textbf{Skepxel distance loss} indicates two-stream training where the Skepxels are used for pair-wise distance loss calculation. \textbf{NoAug} specifies evaluation on the original test samples only.
\renewcommand{\thefigure}{8}
\begin{figure*}[!hbtp]
\centering
\vspace{-0.9em}
\includegraphics[width=\textwidth]{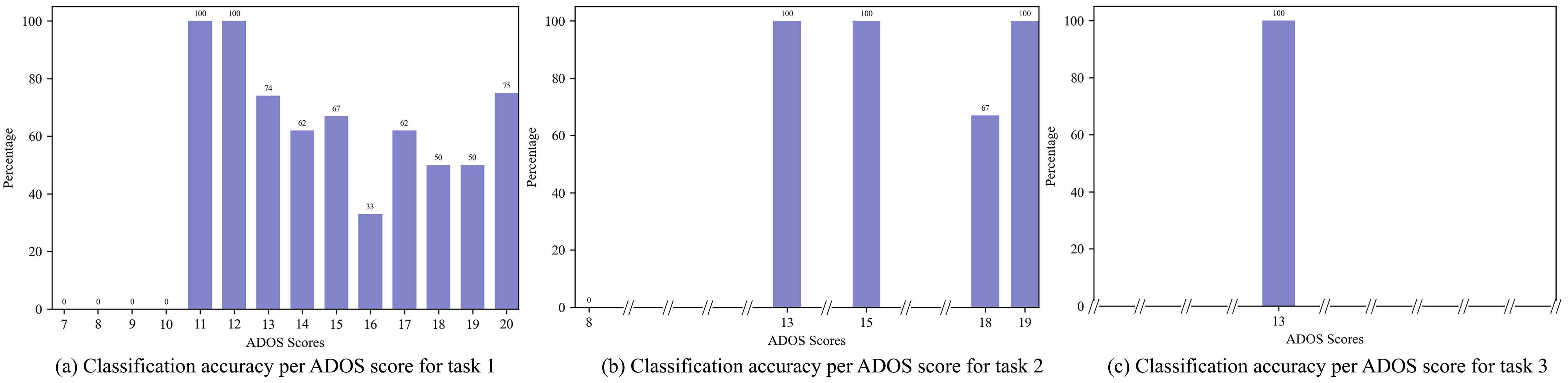}
\vspace{-1.9em}
\caption{Classification accuracy per ADOS scores for individual tasks (a) task 1 - Turn taking (b) task 2 - Imitation and (c) task 3 - Joint attention.} 
\label{all_data_per_task}
\vspace{-0.8em}
\end{figure*}  

\begin{table*}[h]
\centering
\footnotesize
\begin{tabular}{l|cccccccc|}
\cline{2-9}
                                   & \multicolumn{8}{c|}{\textbf{Accuracy}}                                                                                                                                                                                                                                                                                                                                                                                                                                 \\ \cline{2-9} 
                                   & \multicolumn{4}{c|}{\textbf{Block data}}                                                                                                                                                                                                     & \multicolumn{4}{c|}{\textbf{Random data}}                                                                                                                                                                               \\ \cline{2-9} 
                                   & \multicolumn{1}{c|}{MSG3D} & \multicolumn{1}{c|}{\begin{tabular}[c]{@{}c@{}}+Angle\\ infusion\end{tabular}} & \multicolumn{1}{c|}{\begin{tabular}[c]{@{}c@{}}+Skepxel\\ distance loss\end{tabular}} & \multicolumn{1}{c|}{NoAug}          & \multicolumn{1}{c|}{MSG3D} & \multicolumn{1}{c|}{\begin{tabular}[c]{@{}c@{}}+Angle\\ infusion\end{tabular}} & \multicolumn{1}{c|}{\begin{tabular}[c]{@{}c@{}}+Skepxel\\ distance loss\end{tabular}} & NoAug          \\ \hline

\multicolumn{1}{|c|}{\textbf{\makecell{Avg \\$\pm$ SD}}} & \multicolumn{1}{c|}{\makecell{86.38 \\$\pm$ 5.48}}    & \multicolumn{1}{c|}{\textbf{\makecell{89.10 \\$\pm$ 6.67}}}                                            & \multicolumn{1}{c|}{\textbf{\makecell{90.86 \\$\pm$ 4.84}}}                                                   & \multicolumn{1}{c|}{\textbf{\makecell{93.00 \\$\pm$ 4.82}}} & \multicolumn{1}{c|}{\makecell{91.24 \\$\pm$ 3.19}}    & \multicolumn{1}{c|}{\textbf{\makecell{92.66 \\$\pm$ 3.01}}}                                            & \multicolumn{1}{c|}{\textbf{\makecell{91.48 \\$\pm$ 3.90}}}                                                   & \textbf{\makecell{92.33 \\$\pm$ 3.67}} \\ \hline

\end{tabular}
\vspace{-0.8em}
\caption{Average results with standard deviation from ten fold cross-validation on the Gait dataset. The results clearly demonstrate that the inclusion of additional modules in our proposed architecture yields notable improvements in comparison to the baseline model: MSG3D.}
\label{CrossValidation_results}
\vspace{-1em}
\end{table*}

Table~\ref{tab_gait_dataset_results} illustrates the comparison between our proposed method and existing work on the Gait and Full Body Movement dataset. Ahmed et al. \cite{Ahmed_2021} did not share the train-test subject split and presented a single experimental result. We perform more rigorous experiments with 10-fold cross-validation, ensuring that every subject ends up in the test set. Our proposed method achieves 93\% accuracy on average on the 10-folds, with a minimum of 86.67\% and a maximum of 100\% accuracy.\\
\vspace{-1em}
\begin{table}[!hbtp]
\centering
\footnotesize
\begin{tabular}{|c|c|c|c|}
\hline
\textbf{Method}       & \textbf{Data}             & \textbf{Subject selection} & \textbf{Accuracy} \\ \hline
Ahmed et al. \cite{Ahmed_2021}                     & skeleton                  & Random                     & 92.00             \\ \hline
\multirow{4}{*}{Ours} & \multirow{4}{*}{skeleton} & Random (Max)               & \textbf{96.67}    \\ \cline{3-4} 
                      &                           & Random (Avg)               & \textbf{92.33}    \\ \cline{3-4} 
                      &                           & Block (Max)                & \textbf{100.00}   \\ \cline{3-4} 
                      &                           & Block (Avg)                & \textbf{93.00}    \\ \hline
\end{tabular}
\vspace{-0.7em}
\caption{Comparison between our results and existing work on the Gait and Full Body Movement dataset. \label{tab_gait_dataset_results}}
\vspace{-0.4em}
\end{table}

\textbf{Results on DREAM dataset:} 
The DREAM dataset contains only ASD samples with corresponding ADOS scores and ADOS-related information. Samples in this dataset are way longer than the Gait and Full Body Movement dataset, with a maximum number of frames of around 60K. Therefore, we finetune our model on this dataset using samples with one minute duration. Then the trained model is used to extract features from the other frames of all samples. Finally, we use Support Vector Regression (SVR) to predict the ADOS scores. 

We calculate the corresponding classes using the predicted ADOS scores and the corresponding ADOS module and age following the metrics provided in \cite{Katherine2009}. NS class includes the following population: for ADOS module 1 - ages ($\geq$3 and $\leq$6 years) with scores $\leq$10 and for module 2 - ages (3 and 4) with scores (6 and 7) and ages (5 and 6) with scores $\leq$6 \cite{Katherine2009}. ASD class includes: for module 1 - ages $\geq$6 with scores ($>$10 and $\leq$15) and for module 2 - ages (3 and 4) with scores ($>$6 and $\leq$9) and ages (5 and 6) with scores =8 \cite{Katherine2009}. AUT class includes: for module 1 - ages ($\geq$3 and $\leq$6) with scores $>$15 and for module 2 - ages (3 and 4) with scores $>$9 and ages (5 and 6) with scores $>$8 \cite{Katherine2009}. Table~\ref{svr_results_skel_eye} represents the error rates for score regression, spearman correlation (SP), \textit{P} values, and classification accuracy calculated from the predicted scores. Spearman correlation and \textit{P} values indicate that the predicted ADOS scores are highly correlated with the actual scores measured by expert professionals. 

\begin{table}[!hbtp]
\centering
\footnotesize
\begin{tabular}{P{0.01cm}|P{1.5cm}|P{1cm}|P{1cm}|P{1.2cm}|}
\cline{2-5}
                                   & \textbf{Error rate} & \textbf{SP} & \textit{\textbf{P}} & \textbf{Accuracy} \\ \hline
\multicolumn{1}{|l|}{\textbf{\makecell{Avg \\$\pm$ SD}}} & \makecell{2.91 \\$\pm$ 0.27}         & \makecell{0.34 \\$\pm$ 0.07} & \makecell{.002 \\$\pm$ .003}         & \makecell{51.56 \\$\pm$ 5.10}      \\ \hline
\end{tabular}
\vspace{-0.7em}
\caption{SVR Regression for 10-fold cross-validation. Accuracy represents \% of correct classification calculated using predicted ADOS Scores.}
\label{svr_results_skel_eye}
\vspace{-1.5em}
\end{table}

Subjects in the DREAM dataset perform three tasks: Turn taking, Imitation, and Joint attention. We analyze the overall and the per-task accuracy and achieve 78.6\% average accuracy with individual tasks. Figure~\ref{all_data_all_action} illustrates the classification accuracy per ADOS scores for all tasks.

\vspace{0.3em}
\renewcommand{\thefigure}{7}
\noindent%
\begin{minipage}{\columnwidth}
\makebox[\columnwidth]{
  \includegraphics[keepaspectratio=true,scale=0.17]{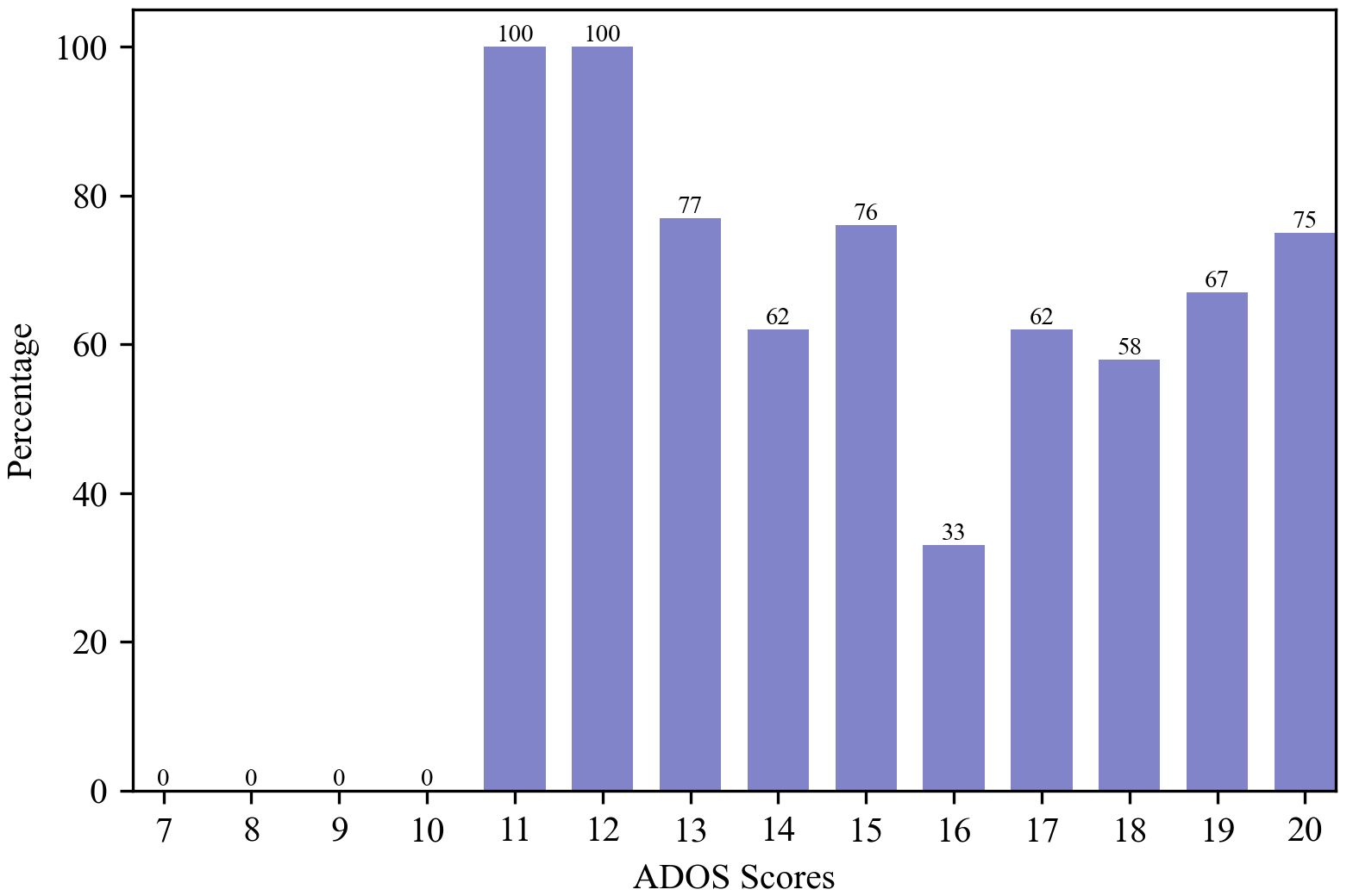}}
  \vspace{-2em}
\captionof{figure}{Classification accuracy per ADOS scores for all tasks.}\label{all_data_all_action}
\end{minipage}

\vspace{1em}
Figure~\ref{all_data_per_task} represents classification accuracy for individual tasks. For tasks 1 and 2, the model fails to correctly classify ADOS total in the range of 7-10. This ADOS score range is the hardest as they fall within the NonSpectrum class, where children do not have autism but exhibit very subtle atypical behaviors.

\section{Discussion}\label{Discussion}
The gait and full body movement dataset comprises single gait cycles, with a restricted number of frames. Due to their brevity, anomalous behavioral expressions exhibited by children with ASD are scant in these sequences. However, our statistical analysis extracted valuable insights into the gait and gesture traits. The observed gait asymmetry and increased joint angle, motion, and distance in standing pose can provide a significant attribute for autism analysis. Longitudinal data with extended duration will provide a better understanding of autistic gesture expressions. It will facilitate faster diagnosis and a more personalized therapy and support system.

Experimental results on the DREAM dataset indicate that our model fails to classify samples with lower ADOS scores ranging from 7 to 10. This range constitutes the non-Spectrum class, where subjects have mild ASD-like behavioral patterns with very few atypical traits. Additionally, the dataset incorporates a limited number of samples with no visible distinction among samples from different classes. On the other hand, the methodologies for measuring ADOS scores are sub-standardized, adding further complexity to the prediction process. We use a range of tolerance values to mitigate this issue while associating predicted ADOS scores with classes. Despite these challenges, our proposed methodology demonstrates superior performance, particularly for higher ADOS scores within the ASD and AUT classes. These two classes show more physical and behavioral anomalies and require substantial assistance and support systems. Our proposed method offers a feasible assessment solution for such cases. 

Moreover, privacy concerns have significantly impeded progress in autism research. Employing a skeleton video-based assessment system can alleviate these concerns and enable a more comprehensive approach to autism analysis. Our present study opens a new research area to anonymize and automate the tedious process of autism diagnosis and ADOS score prediction.
\section{Conclusion}\label{Conclusion}
This paper presents a comprehensive analysis of autism spectrum disorder using gait and gesture-based approaches from skeleton videos. Our statistical analysis suggests that children with ASD display asymmetrical gait patterns and higher mean joint angle distributions. These findings align with the overall trend in atypical behavioral patterns observed in individuals with ASD. Our proposed early angle embedding technique enhances GCN performance by emphasizing atypical gesture patterns with greater precision to extract robust spatio-temporal features. Moreover, we leverage the concept of ``Skepxels" to enable multi-modal training without the need for supplementary input modalities. Our experimental results indicate that utilizing skeleton data holds significant potential for advancing our understanding of autism related behaviours and could serve as a promising avenue for future research.

\section{Acknowledgement}
Professor Ajmal Mian is the recipient of an Australian Research Council Future Fellowship Award (project number FT210100268) funded by the Australian Government. Sania Zahan is the recipient of a University Postgraduate Award and University of Western Australia International Fee Scholarship.


{\small
\bibliographystyle{ieee_fullname}
\bibliography{egbib}
}

\end{document}